\documentclass[conference,a4paper]{IEEEtran}
\usepackage{xcolor}
\usepackage{balance}
\usepackage{amsmath}
\usepackage{subfigure}
\usepackage{booktabs}
\usepackage{tabularx}

\ifCLASSINFOpdf
  \usepackage[pdftex]{graphicx}
\else
  \usepackage[dvips]{graphicx}
\fi
\ifCLASSOPTIONcompsoc
 \usepackage[caption=false,font=normalsize,labelfont=sf,textfont=sf]{subfig}
\else
 \usepackage[caption=false,font=footnotesize]{subfig}
\fi
\begin{document}
%
\title{Large Language Model-Based Intelligent Antenna Design System}

\author{\IEEEauthorblockN{
Tao Wu\IEEEauthorrefmark{1},   
Kexue Fu\IEEEauthorrefmark{2},   
Qiang Hua\IEEEauthorrefmark{3},    
Xinxin Liu\IEEEauthorrefmark{1},
Bo Liu\IEEEauthorrefmark{1}
}                                     
\IEEEauthorblockA{\IEEEauthorrefmark{1}
James Watt School of Engineering, University of Glasgow, G12 8QQ Glasgow, U.K.}
\IEEEauthorblockA{\IEEEauthorrefmark{1}
\{t.wu.1,x.liu.11\}@research.gla.ac.uk, bo.liu@glasgow.ac.uk}
\IEEEauthorblockA{\IEEEauthorrefmark{2}
Department of Electrical Engineering, City University of Hong Kong, Hong Kong SAR., kexuefu2-c@my.cityu.edu.hk}
\IEEEauthorblockA{\IEEEauthorrefmark{3}
Department of Engineering and Technology, University of Huddersfield, HD1 3DH Huddersfield, U.K., q.hua@hud.ac.uk}  
}



\maketitle

\begin{abstract}
Antenna simulation typically involves modeling and optimization, which are time-consuming and labor‑intensive, slowing down antenna analysis and design. This paper presents a prototype of a large language model (LLM)-based antenna design system (LADS) to assist in antenna simulation. LADS generates antenna models with textual descriptions and images extracted from academic papers, patents, and technical reports (either one or multiple), and it interacts with engineers to iteratively refine the designs. After that, LADS configures and runs an optimizer to meet the design specifications. The effectiveness of LADS is demonstrated by a monopole slotted antenna generated from images and descriptions from the literature. To improve gain stability across the 3.1–10.6 GHz ultra-wide band, LADS modifies the cross-slot into an H-slot and changes substrate material, followed by parameter optimization. As a result, the gain variation is reduced while maintaining the same gain level. The LLM-enabled antenna modeling (LEAM) is available at: https://github.com/TaoWu974/LEAM.
\end{abstract}

\vskip0.5\baselineskip

\section{Introduction}
In antenna research and applications, there is a demanding need for exploring novel antenna structures and reproducing designs from the literature. Antenna modeling and optimization are often unavoidable in this process, costing a lot of time and effort. Depending on the user’s experience, modeling and optimizing a complex antenna could take from dozens of hours to days, and the demand for such simulations continues to rise with the growth of the antenna industry. Therefore, it is desirable to reduce repetitive and error-prone modeling work and employ the human-free optimization. This motivates the automation of antenna modeling and optimization, speeding up the antenna design and innovation process.

It is common practice to use the graphical user interface (GUI) in electromagnetic (EM) simulation tools for antenna modeling and visualization first, followed by implementing external optimization scripts, typically in Python or MATLAB, to control the simulation and tune the design parameters. The visualized antenna structures can be represented in macro files, which are command lines describing the model. Macros also act as an interface between external optimization scripts and EM simulation tools to enable automated control of the simulations. For example, CST Microwave Studio employs Visual Basic for Applications (VBA) macros, and HFSS supports IronPython macros. To enable automation, two types of macro files are typically required: one for constructing the antenna model and another for controlling the simulation flow and parameter updates during optimization.

However, modeling macro files are complex, and manual scripting is even more time-consuming than drawing the model using GUI. For example, importing material to an object is a straightforward click using the GUI of EM simulation tools, but tens of parameters are needed in the macro file. The gap between the simple language description of the user and the complex macro file that can correctly define the antenna model becomes the bottleneck, which has not been solved till now. Moreover, even with existing optimization scripts, it remains nontrivial for antenna engineers to configure optimization, including setting optimizer parameters and preparing macros to link simulation tools and optimization scripts. For example, design variables and their search ranges must be defined correctly considering geometry constraints, while from a view of optimization goals, constraints and objectives need to be handled separately. These challenges make it difficult for designers to obtain optimal parameters efficiently, and the misconfigurations can easily lead to optimization failures.

A recent innovation, Large Language Models (LLMs), provides a promising solution, bridging prompt and codes from macros and optimization scripts by its ability of geometry understanding \cite{parra2024can}, parameterization \cite{wu2023cad} and code generation \cite{lin2024llm}. Prompt is the input to the LLMs and can be texts and/or images. Although have not been used in antenna design, LLMs are investigated for automated macro scripting for 3D mechanical structure modeling. Some pioneer research works and findings are as follows. \cite{zamfirescu2023johnny} shows weak prompts can lead to poor responses and task failures. \cite{li2025llm4cad} shows that LLMs with vision capabilities (i.e., image recognition) outperform text-only LLMs as the modeling complexity increases. \cite{badagabettu2024query2cad} collects macro templates and benchmark test cases to assess LLM-based modeling techniques. SolidWorks employs LLMs in modeling mechanical components \cite{deng2024investigation}.

Despite the advancements in applying LLMs to 3D modeling, three limitations are restricting their applications in antenna modeling and optimization: 
\begin{itemize}
    \item There is limited research considering multi-modal input (i.e., not only textual descriptions, but also image input) \cite{li2025llm4cad}, but is crucial for LLM-enabled antenna modeling, especially for reproducing models from the literature.
    \item Antenna modeling typically involves creating multiple solids and performing Boolean operations among them, which differs from the mechanical modeling process of selecting one model and configuring its parameters \cite{li2025llm4cad}.
    \item To date, there has been little research aimed at integrating antenna modeling and optimization into a unified workflow.
\end{itemize}

To address these challenges, the \underline{L}LM-based \underline{a}ntenna \underline{d}esign \underline{s}ystem (LADS) is proposed, which consists of an \underline{L}LM-\underline{e}nabled \underline{a}ntenna \underline{m}odeling method (LEAM) and LLM-controlled optimization. The modeling macro scripting process is divided into systematic steps, and a set of LLM tools based on OpenAI LLM Application Programming Interface (APIs) with designated prompts including responses from other LLM tools are designed for these steps. The input of LEAM is textual descriptions and images (if available), and the output is the antenna model. Currently, CST Microwave Studio is used as the EM simulator. To the best of our knowledge, this is the first study to investigate LLM-enabled antenna automated modeling. After that, human engineers can propose new design goals to LLM, then review and select solutions from LLMs to refine the antenna design. Finally, LLM configures and runs parameter optimization using SB-SADEA, a self-adaptive Bayesian neural network surrogate-model-assisted differential evolution for antenna optimization \cite{liu2022efficient}.

\section{The System Overview and Methodology}
\subsection{The LADS workflow}
\begin{figure}[!h]
\centerline{\includegraphics[scale=0.6]{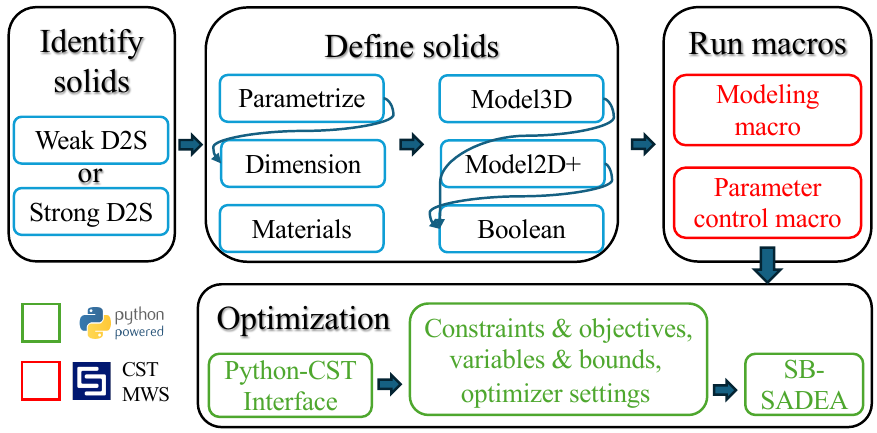}}
\caption{LEAM workflow}
\label{fig:workflow}
\end{figure}

In LADS, the antenna modeling and optimization are divided into systematic steps, and for each step, several tools are created. The tools are shown in Fig. \ref{fig:workflow} and Table \ref{tab:tools}. Users typically employ the tools sequentially, except for the first step, there is selection, which will be detailed later.

\begin{table}[ht]
    \centering
    \caption{LEAM Tools with I/O, and LLM Model}
    \begin{tabular}{l l l l} 
        \hline
        \textbf{Tool} & \textbf{Input} & \textbf{Output} & \textbf{LLM} \\
        \hline
        Weak D2S      & Text, Image (if available)          & Solid\_List.txt & o1    \\
        Strong D2S   & Text, Image (if available)     & Solid\_List.txt & o1    \\
        Parameterize   & Text, Image (if available)     & Para.bas        & o1/4o    \\
        Dimension   & Text, Image (if available)     & Solids\_Dims.txt& o1/4o   \\
        Materials   & Solid\_List.txt   & Materials.bas   & 4o  \\
        Model3D     & Solids\_Dims.txt \& *.bas & 3D\_Model.bas & o1  \\
        Model2D+   & Solids\_Dims.txt \& *.bas & 2D+\_Model.bas& o1 \\
        Boolean     & Solids\_Dims.txt \& *.bas & Boolean.bas   & o1 \\
        Modeling      & *.bas            & Model.cst       & N/A\\
        ParaControl      &  *.txt \& Para.bas           & Simu.bas     & o1\\
        Interface      & Simu.bas         & Interface.yml       & o1\\
        ConfigOpt      & Text \& Interface.yml            & Opt\_config.yml       & o1\\
        SB-SADEA      & *.yml            & Optimal.cst       & N/A\\
        \hline
    \end{tabular}
    \label{tab:tools}
\end{table}

As shown in Fig. \ref{fig:workflow}, the design process is divided into four stages, which are identifying solids, defining solids, executing macros, and optimization. (In this work, CST Microwave Studio (CST MWS) is used, where solid is the fundamental modeling unit.) In the first stage, if the provided description is sufficient, the strong descriptions to solids (Strong D2S) tool will be used; otherwise, the weak descriptions to solids (Weak D2S) tool will be used. Both tools have the same output, which lists all solids with their materials, types (e.g., substrate, feeding line, slot, etc.), geometric shapes (e.g., brick, cylinder, extruded 2D shape, etc.), and positions. The Strong D2S tends to translate the user's instructions only, while Weak D2S autofills the incomplete descriptions with LLM's knowledge about antennas.

The second stage is to define solids. Using the solid list obtained from the first stage, the Parameterize tool is then employed to define parameters and generate a Para.bas file to import parameters into the solids. After that, with defined parameters and descriptions (with images if available), the solid list is updated by adding detailed dimensions and positions. Next, the Materials tool queries the EM modeling tool’s material library to retrieve the materials specified in the solid list and subsequently generates the Materials.bas to import these materials into the solids. 

After that, 3D and 2D+ solids modeling macros are generated separately, followed by boolean operations. Given the solid list with dimensions, parameters, and materials, the Model3D tool identifies and models the 3D shapes, and produces the 3D\_Model.bas file. 2D+ shape refers to extruded or rotated 2D shapes. The Model2D+ tool first models closed 2D shapes and then extrudes and/or rotates these 2D shapes into 3D shapes, and produces the 2D+\_Model.bas file. Then, given the modeled solids in the above bas files and the solid list with dimensions, the Boolean tool determines sequential boolean operations in the Boolean.bas file. For example, \textit{adding} the connected solids, \textit{subtracting} the slot solids.

The third stage is to execute macro files. The Modeling tool executes all bas files obtained previously, including Para.bas, Materials.bas, 3D\_Model.bas, 2D+\_Model.bas, and Boolean.bas. These macros are saved to the history list of CST Microwave Studio. The automated antenna modeling is completed. After that, ParaControl tool determines the parameters to be optimized and generates the corresponding parameter control and simulation bas file Simu.bas. 

\begin{figure*}[t]
\centering
\includegraphics[scale=0.6]{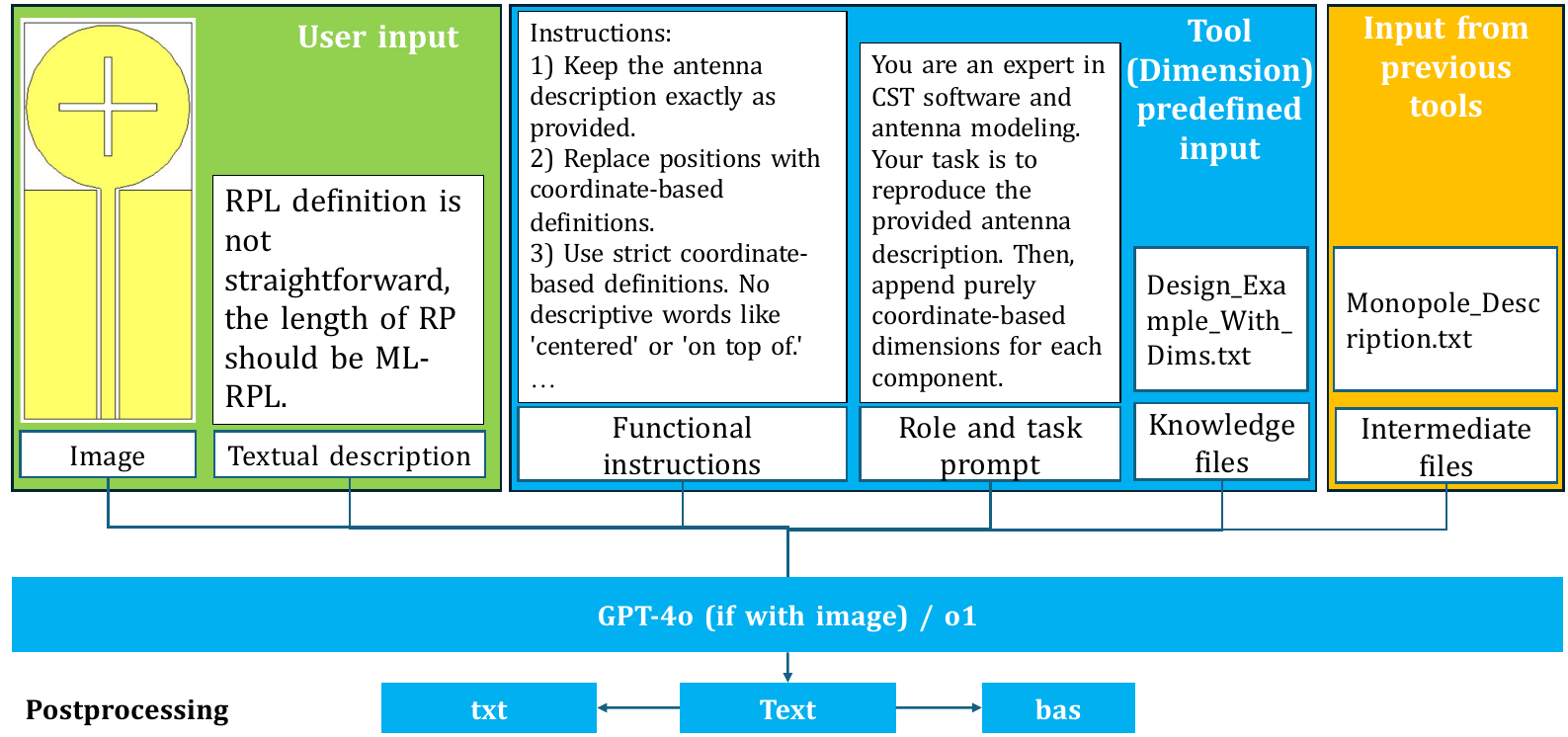}
\caption{The LLM-based Dimension tool I/O}
\label{fig:tool_io}
\end{figure*}

Finally, the Interface tool reads in the Simu.bas file to generate an interface.yml file to configure the interface of controlling the design variables and calling CST MWS to run EM simulations, while the ConfigOpt tool creates an Opt\_config.yml file to configure the optimization, including bounds, constrains, objectives and other optimization settings. Finally, SB-SADEA \cite{liu2022efficient} runs the optimization of antenna parameters with the given *.yml files, and outputs an optimal design.

\subsection{LLM tool creation}
As said above, bas executing tool, SB-SADEA optimizer tool and 11 LLM tools are defined in LADS. Fig. \ref{fig:tool_io} shows a tool structure using the Dimension tool as an example. Each tool accepts pre-defined input, input from former tools, and user input. The pre-defined input is the major effort to link LLM with antenna modeling, including role and task prompts, functional instructions, and knowledge files. The role and task prompts provide a task-oriented identity to help the LLM align with task requirements \cite{zheng2023helpful}. The functional instructions define the task details and output rules (e.g., \textit{"Output VBA code only"}) \cite{zhang2024instruct}. Knowledge files are examples of macros and formatted output to guide the LLM.

It is worth noting that various LLM models are carefully selected for each tool. With the advancement of reasoning LLMs such as OpenAI o1 \cite{jaech2024openai} and Deepseek R1 \cite{guo2025deepseek}, LLMs can now build more complex workflows using the Chain of Thought (CoT) \cite{jaech2024openai}, aligning seamlessly with complex scripting requirements of antenna modeling. Our pilot experiments show that OpenAI o1 is better at understanding a topology and dismantling it into basic geometric shapes, while GPT-4o outperforms OpenAI o1 in image recognition. Hence, to trade off the reasoning ability and image recognition ability, the general principle is that the base model (GPT-4o-2024-11-20) is used for most multi-modal input and the reasoning model (o1-2024-12-17) is used for text-only input. However, when advanced reasoning is critical, such as the D2S tool, OpenAI o1 is employed even though the input is multi-modal.

\subsection{Demonstration of LEAM by a simple case}
\begin{figure}[h!]
\centerline{\includegraphics[scale=0.35]{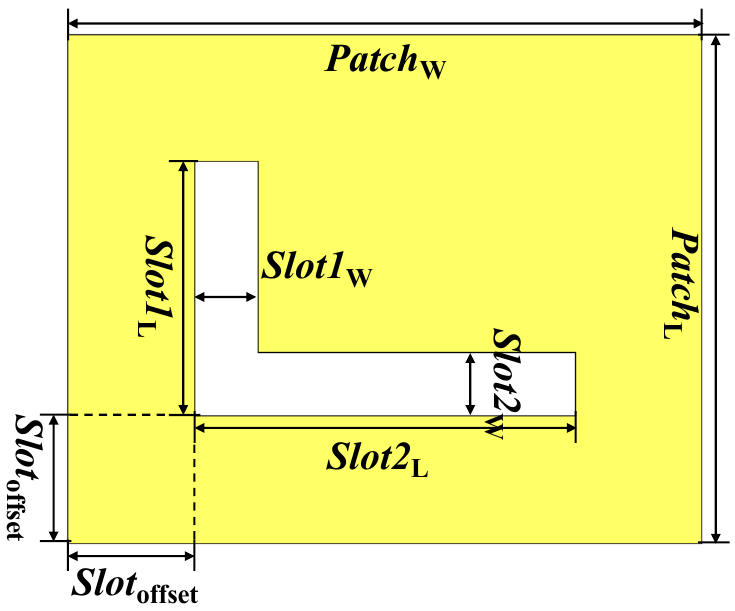}}
\caption{Example: L-slot on the patch. $Patch_{L}$ = 8, $Patch_{W}$ = 10, $Slot1_{L}$ = 4, $Slot1_{W}$ = 1, $Slot2_{L}$ = 6, $Slot2_{W}$ = 1, $Slot1_{offset}$ = 2. (unit: mm.)}
\label{fig: L-slot}
\end{figure}

To demonstrate the use of LEAM, a rectangular patch with an L-shape slot example (Fig. \ref{fig: L-slot}) is used.

In the first stage, user descriptions are input. For example, \textit{"Draw an L-shaped 2D polygon in the XY-plane with the desired corner coordinates (e.g., starting at (LX1, LY1), turning at (LX2, LY2), etc.). Close the polygon and extrude in the Z direction."}. The outcome of this stage is a solid list including two solids, which are a brick for the patch and an extruded 2D shape for the slot.

In the second stage, a solid list with dimensions and two macro files are generated: Para.bas to import 7 parameters (\textit{$Patch_W$, $Patch_L$, $Slot1_W$, $Slot1_L$, $Slot2_W$, $Slot2_L$, $Slot_{Offset}$}) and Materials.bas to import copper (pure) as the patch's material. The position of the brick for the patch is defined by the coordinate ranges (\textit{"Xmin=0, Xmax=PatchW, Ymin=0, Ymax=PatchL, Zmin=0, Zmax=0.035"}), while the L-shape slot is defined by six points forming the L-shape polygon whose Z range is of [0, 0.035]. The descriptions are as follows.
\begin{description}
\item Start from [$\mathrm{Slot_{Offset}}$,$\mathrm{Slot_{Offset}}$] \\
To [$\mathrm{Slot_{Offset} + Slot2_L}$, $\mathrm{Slot_{Offset}}$] \\
To [$\mathrm{Slot_{Offset} + Slot2_L}$, $\mathrm{Slot_{Offset} + Slot2_W}$] \\
To [$\mathrm{Slot_{Offset} + Slot1_W}$, $\mathrm{Slot_{Offset} + Slot2_W}$] \\
To [$\mathrm{Slot_{Offset} + Slot1_W}$, $\mathrm{Slot_{Offset} + Slot1_L}$] \\
To [$\mathrm{Slot_{Offset}}$, $\mathrm{Slot_{Offset} + Slot1_L}$] \\
To [$\mathrm{Slot_{Offset}}$,$\mathrm{Slot_{Offset}}$]
\end{description}

In the third stage,  a 3D\_Model.bas for the patch, a 2D+\_Model.bas for the L-shape slot, and a Boolean.bas for subtracting the L-shape slot from the patch are generated. To model the 2D+ L-shape slot solid, 2D+\_Model.bas first creates an L-shape polygon and then extrudes it to a height of 0.035 (the same as the patch). The fourth stage executes the macro files and the resulting antenna model is in Fig. \ref{fig: L-slot}.

\section{Experimental Results}

In this section, an ultra-wide band (UWB) slotted monopole antenna's modeling and optimization are used to demonstrate LADS, and some other cases of modeling method LEAM can accessed by https://github.com/TaoWu974/LEAM. The slotted patch antenna model is reproduced using descriptions and images scanned from \cite{liu2022efficient}.

\begin{figure}[h]
    \centering
    \subfigure[Descriptions]{%
        \includegraphics[scale=0.52]{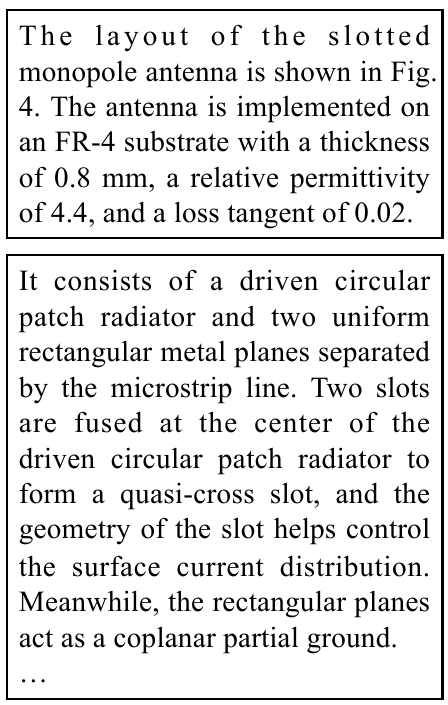}%
    }%
    \hfil
    \subfigure[Topology]{%
        \includegraphics[scale=0.48]{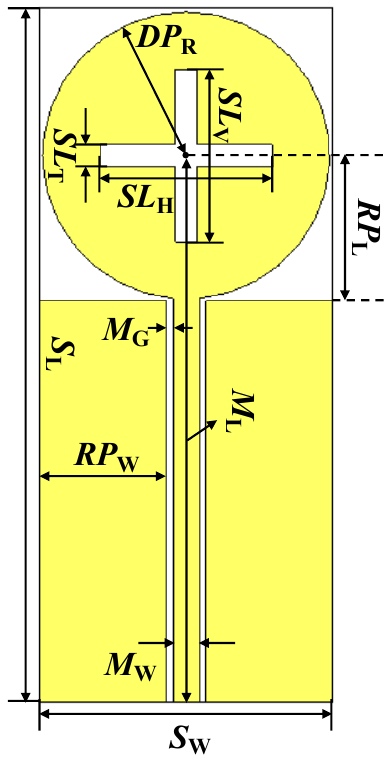}%
    }%
    \caption{Cross-slotted monopole antenna, retrieved from Section IV of \cite{liu2022efficient}.}
    \label{fig:combined_monopole}
\end{figure}

The description, topology figure, and parameter list figure are scanned from \cite{liu2022efficient} (Fig. \ref{fig:combined_monopole}). The strong D2S tool identifies seven 3D solids (substrate, circular patch, horizontal slot, vertical slot, left ground plane, and right ground plane) with 12 parameters (10 parameters from the paper's parameter list and 2 thickness parameters in the text) and 2 materials, FR-4 for the substrate and copper for the patch. A typo in \cite{liu2022efficient} is corrected by adding a prompt to the Parameter tool ($S_L = M_L + DP_R + 0.2$). Then, to refine the gain stability over the frequency band from 3.1 GHz to 10.6 GHz, we pass this requirement, identified solids, and descriptions from the paper as a prompt to the LLM. Ten solutions are given, and two of them are chosen by the engineer: 1) Change the substrate material from FR-4 to Rogers RT 5880, 2) Change the slot structure from the cross-slot to the H-slot. Some other unselected solutions include adding parasitic element, increasing substrate thickness, and applying a taper feeding.

\begin{table}[htbp]
  \centering
  \small
  \caption{Design Variables of the H-Slotted Monopole Antenna}
  \label{tab:design_variables}
  \renewcommand{\arraystretch}{1.0}
  \setlength{\tabcolsep}{4pt}
  \begin{tabularx}{\linewidth}{@{}>{\raggedright\arraybackslash}X >{\raggedright\arraybackslash}X@{}}
    \toprule
    \textbf{Design Variable} & \textbf{Design Variable} \\
    \midrule
    Circle patch radius ($\mathrm{DP}_R$)        & Substrate width ($\mathrm{S}_W$) \\
    Horizontal slot width ($\mathrm{SW}_H$)      & Vertical slot width ($\mathrm{SW}_V$) \\
    Horizontal slot length ($\mathrm{SL}_H$)     & Vertical slot length ($\mathrm{SL}_V$) \\
    Vertical slot offset ($\mathrm{S}_O$)        & Ground plane length ($\mathrm{G}_L$) \\
    Microstrip width ($\mathrm{M}_W$)            & Microstrip gap ($\mathrm{M}_G$) \\
    Feed guide width ($\mathrm{F}_W$)            & Microstrip length ($\mathrm{M}_L$) \\
    \midrule
    \multicolumn{2}{@{}l@{}}{\textbf{Substrate length ($\mathrm{S}_L$)} = $\mathrm{M}_L + \mathrm{DP}_R + 0.2$ (mm)}\\
    \multicolumn{2}{@{}l@{}}{\textbf{Ground plane width ($\mathrm{G}_W$)} = $(\mathrm{S}_W - 2\times\mathrm{M}_G - \mathrm{M}_W)/2$ (mm)}\\
    \bottomrule
  \end{tabularx}
\end{table}

\begin{figure}
    \centering
    \subfigure{%
        \includegraphics[scale=0.5]{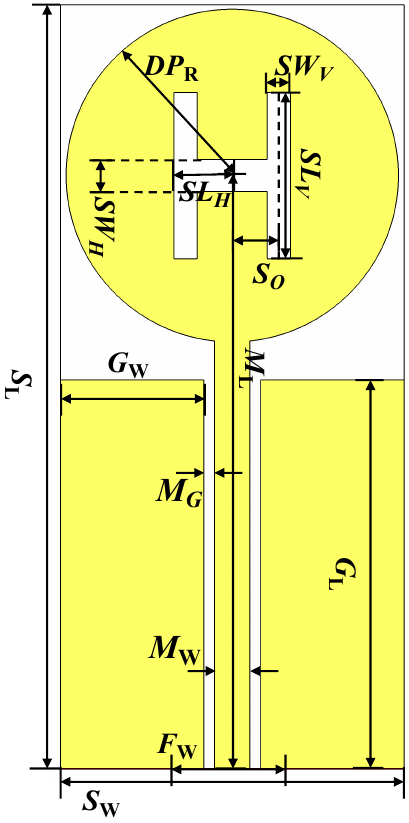}%
    }%
    \caption{H-slotted monopole antenna}
    \label{fig:H_Slotted_antenna}
\end{figure}

As a result, the modeled antenna is shown in Fig. \ref{fig:H_Slotted_antenna}, and 12 design variables are summarized in Table \ref{tab:design_variables}. The antenna's substrate is of Rogers RT 5880 with a thickness of 0.8 mm. It is noteworthy that the solution space of the H-slot topology encompasses that of the cross-slot structure. Specifically, when the horizontal offset between the two vertical slots is set to zero, or when their widths exceed the offset, the H-slot degenerates into a cross-slot configuration. Subsequently, SB-SADEA is employed to optimize these variables. The antenna is discretized with a mesh density of 15 cells per wavelength, resulting in approximately 75{,}000 mesh cells in total. Each EM simulation takes about 50 seconds to complete.

The optimization aims to minimize the variance of the realized gain ($G$) over the frequency range of 3.1–10.6~GHz, subject to two constraints: (1) the maximum reflection coefficient must remain below $-10$~dB, and (2) the minimum realized gain must exceed 1~dB. The optimization problem is formulated as follows:

\begin{equation}
\begin{aligned}
\min \quad & \mathrm{Var}(G) \\
\text{s.t.} \quad & \max\left(|S_{11}|\right) < -10~\mathrm{dB} \\
& \min(G) > 1~\mathrm{dB}, \quad \forall f \in [3.1,\, 10.6]~\mathrm{GHz}
\end{aligned}
\label{eqn:objective}
\end{equation}

\begin{figure}
    \centering
    \subfigure[Reflection coefficient]{%
        \includegraphics[scale=0.44]{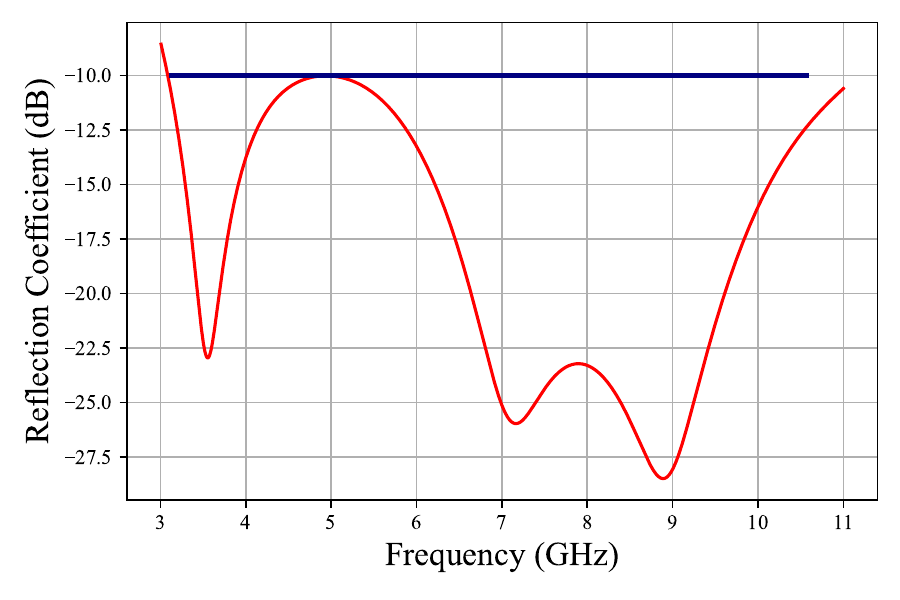}%
    }%
    \hfil
    \subfigure[Realized gain]{%
        \includegraphics[scale=0.44]{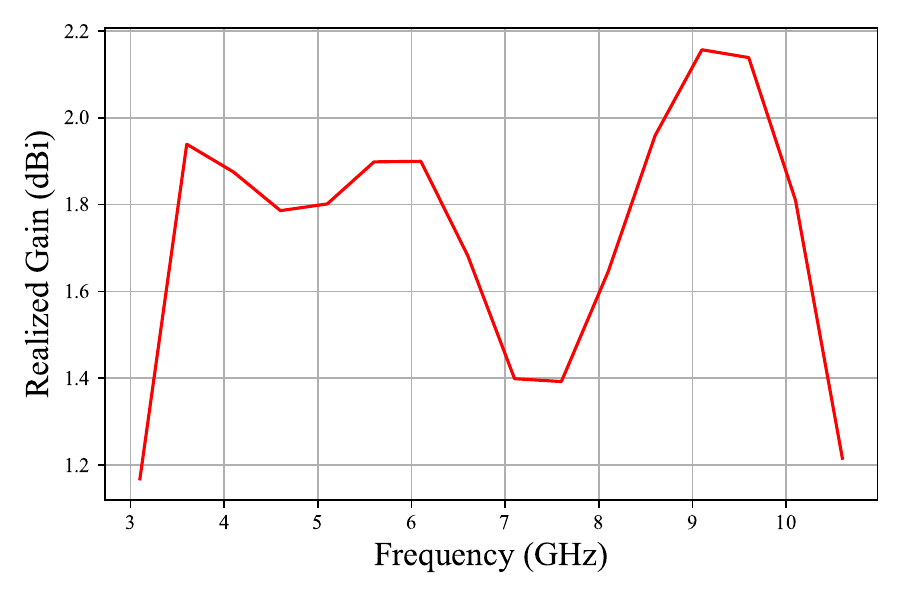}%
    }%
    \caption{The simulated results for the H-slotted antenna}
    \label{fig:Perfs}
\end{figure}

\begin{table}
  \centering
  \small
  \caption{Comparison of Proposed and Existing Antennas}
  \label{tab:comparison}
  \begin{tabular}{@{}lcccc@{}}
    \toprule
    Ref. & $\epsilon_{r}$ & Freq. (GHz) & Gain (dBi) & Volume $\lambda_0 \times \lambda_0 \times \lambda_0$ \\
    \midrule
    *                                  & 2.1 & 3.1-10.6   & 1.17-2.16  & 0.355$\times$0.155$\times$0.008   \\
    \cite{liu2022efficient}            & 4.4 & 3.1-10.6   & 1.19--2.90 & 0.346$\times$0.149$\times$0.008 \\
    \cite{danjuma2020design}            & 4.3 & 3.1-10.6     & 2.17-4.74  & 0.342$\times$0.153$\times$0.008 \\
    \cite{saleem2023ultra}            & 2.2 & 3.2-20.0     & 1-5.36  & 0.361$\times$0.258$\times$0.008 \\
    \cite{khan2022ultra}            & 4.4 & 3.0-12.7     & 0.6-3.6  & 0.310$\times$0.248$\times$0.008 \\
    \cite{elabd2024design}            & 2.2 & 2.05-14.5     & -2-6  & 0.465$\times$0.393$\times$0.018 \\
    \bottomrule
    \addlinespace
    \multicolumn{5}{@{}l@{}}{\footnotesize{* This work}}\\
  \end{tabular}
\end{table}

A design satisfying the two constraints is obtained after 256 EM simulations, followed by continuing optimization and obtaining the optimal design with 721 EM simulations using about 12 hours. Fig. \ref{fig:Perfs} shows the antenna's reflection coefficient and realized gain. Also, the obtained antenna is compared to existing UWB antennas in the literature in Table \ref{tab:comparison}, where $\lambda_0$ is the free-space wavelength at 3.1 GHz (96.8 mm). The realized gain range is reduced to $0.99\textit{ dBi}$ from $1.71\textit{ dBi}$ of the reference \cite{liu2022efficient}, showing improved gain stability.

\section{Conclusion}
In this paper, we present an LLM-based intelligent antenna design system, including an open-source LLM-enabled antenna modeling (LEAM) tool and an LLM-controlled optimization tool. Using an example of modeling and optimizing an antenna from the literature, the effectiveness of LADS has been shown. The effectiveness comes from the proposed systematic framework and the defined tools, including the LLM tools and the optimization tool SB-SADEA in \cite{liu2022efficient}. To the best of our knowledge, LADS is the first attempt to jointly apply LLMs for antenna modeling and optimization. Future works include improving LADS's compatibility with more LLMs (e.g., DeepSeek, Llama) and EM simulation software (e.g., Ansys HFSS, MATLAB Antenna Toolbox). In addition, fine-tuning LLMs on domain-specific data could further enhance the LADS's usability in complex antenna design scenarios.



\bibliographystyle{IEEEtran}
\bibliography{reference}
%









\end{document}